\documentclass[letterpaper,twocolumn]{article}
\usepackage[preprint]{2027}
\usepackage[hyphens]{url}
\usepackage{graphicx}
\usepackage{natbib}
\usepackage{booktabs}
\usepackage{array}
\usepackage{amsmath}
\title{MLIE-CoT: Chain-of-Thought Reasoning via Multi-Level Image Editing}
\author{Baichuan Yang\textsuperscript{1,2}, Dongyuan Xu\textsuperscript{1,2},
  Zhao Yang\textsuperscript{1}, Yong Liu\textsuperscript{1}, Bing Su\textsuperscript{1,*}}
\affiliations{%
  \textsuperscript{1}Renmin University of China, Beijing \qquad
  \textsuperscript{2}Zhongguancun Academy\\
  {\small yang031119@outlook.com, s-xdy26@bza.edu.cn, yangyz1230@ruc.edu.cn, liuyonggsai@ruc.edu.cn}\\
  {\small \textsuperscript{*}Corresponding author: Bing Su (bingsu@ruc.edu.cn)}\\
  Code: \pdfstartlink attr {/Border [0 0 0]} user
  {/Subtype /Link /A << /S /URI /URI (https://github.com/yang031119/MLIE-CoT) >>}%
  \url{https://github.com/yang031119/MLIE-CoT}\pdfendlink}

\begin{document}
\maketitle
\pagestyle{plain}
\thispagestyle{plain}

\begin{abstract}
Chain-of-thought (CoT) reasoning has proven effective in improving the reasoning abilities of large language models by breaking complex tasks into step-by-step solutions. In recent years, researchers have attempted to extend CoT reasoning to visual generative models to improve their ability to handle complex tasks. However, existing CoT methods for text-to-image generation still rely mainly on textual reasoning and therefore cannot fully use and preserve the correct semantic information already present in an image, as even the most detailed language cannot capture all visual information. With advances in image generation, models such as FLUX and BAGEL can generate images conditioned on visual inputs. Recent work has attempted to carry images forward as context in the chain of thought, but this introduces another limitation: unsuitable earlier images can strongly interfere with the model, while repeated whole-image regeneration introduces new hallucinations at each round and causes them to accumulate. To address these issues, we propose MLIE-CoT, a visual chain-of-thought framework based on multi-level image editing. MLIE-CoT learns to select the repair granularity from three levels, preserving correct semantics whenever possible during reasoning while reducing the effect of incorrect content. To improve the ability of multimodal models to route errors of different severity to appropriate repair levels, we construct an image-repair dataset covering different error cases and use it to train a repair-level router. Extensive experiments demonstrate that MLIE-CoT significantly outperforms both its base models and methods using fixed repair strategies, producing clear improvements across multiple benchmarks.
\end{abstract}

\section{Introduction}

High-quality text-to-image models can now generate visually appealing images, yet complex compositional prompts continue to reveal recurring reliability issues: target objects may be missing, colors may be assigned to the wrong objects, counts may differ from the request, and spatial relations may be reversed \citep{chefer2023attendandexcite,ghosh2023geneval,huang2023t2icompbench,wan2025compalign}. Even state-of-the-art diffusion-based models struggle to satisfy every requirement of a complex prompt in a single forward generation, and the difficulty grows as prompts combine more subjects, attributes, counts, and spatial relations \citep{huang2023t2icompbench,wan2025compalign}. Although most constraints may be met, small local errors often remain. Simply emphasizing a previously missed detail does not guarantee a fix, because a new sample may trade the original error for a different one.

\begin{figure}[t]
  \centering
  \includegraphics[width=\columnwidth]{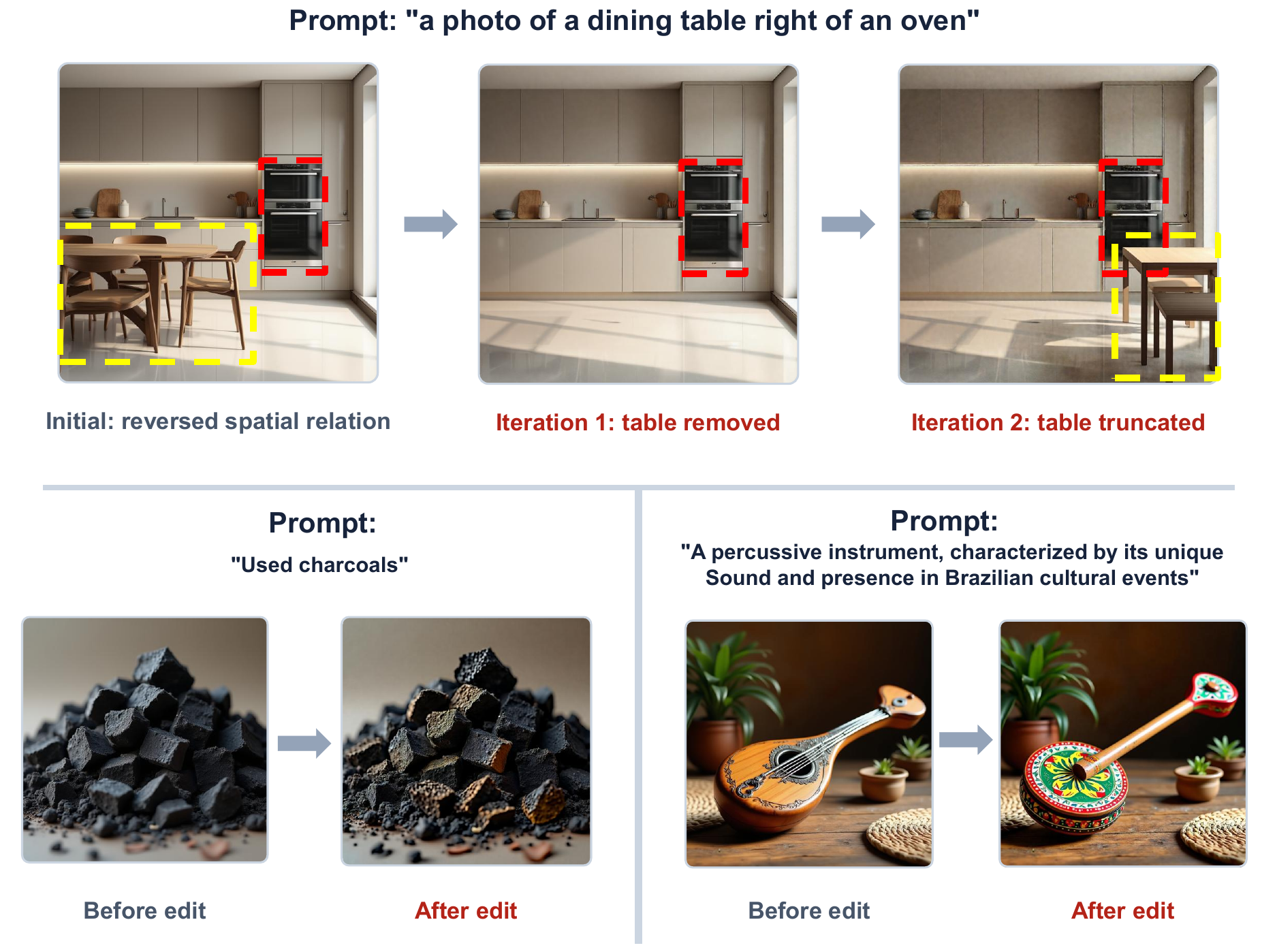}
  \caption{Failure cases caused by negative visual anchors and collateral changes during image-conditioned repair.}
  \label{fig:negative_anchor_collateral_damage}
\end{figure}

\begin{figure*}[t]
  \centering
  \includegraphics[width=0.98\textwidth]{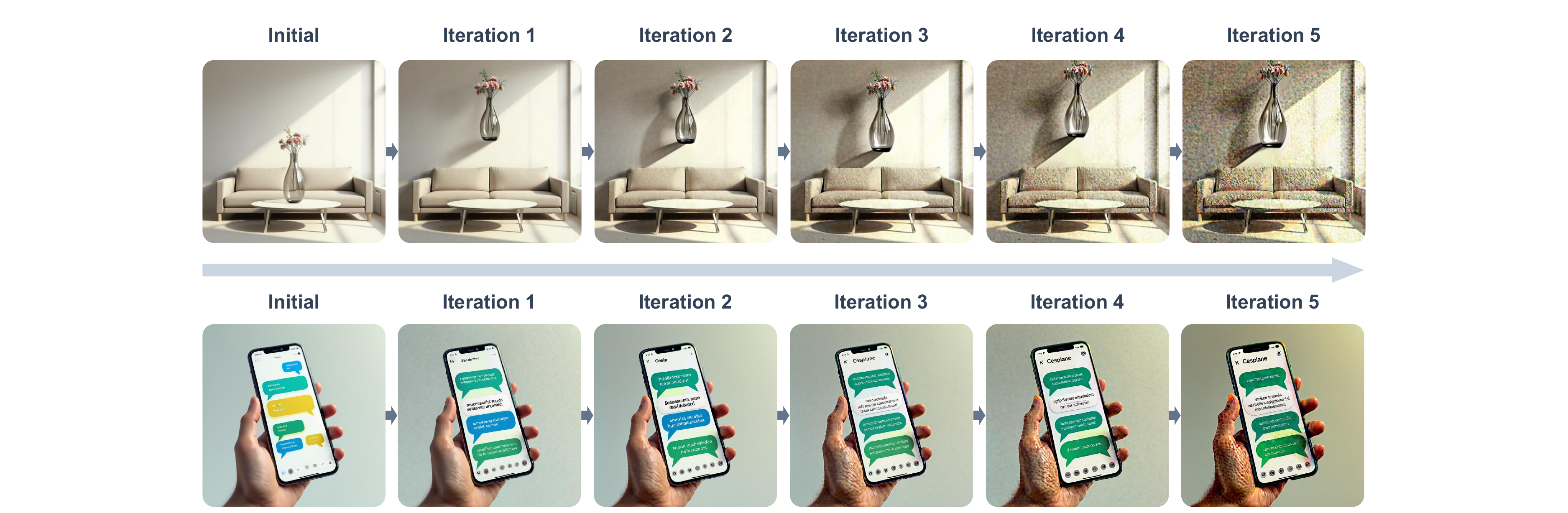}
  \caption{Quality degradation caused by repeated whole-image regeneration. Across both sequences, successive corrections introduce artifacts and overwrite content that was initially correct.}
  \label{fig:unicot_failure}
\end{figure*}

Extending chain-of-thought reasoning to image generation is therefore a natural direction. Visual CoT Diffusion, ImageGen-CoT, and stepwise verification or reinforcement methods have demonstrated the value of intermediate reasoning for image generation \citep{voleti2023visualcotdiffusion,liao2025imagegencot,guo2025imagecot}. Uni-CoT further models macro- and micro-level CoT within a unified understanding-and-generation model, placing text, visual states, and generative actions in a shared model space \citep{qin2025unicot}. Existing visual CoT methods, however, often still reduce reflection to a longer prompt, a layout condition, or another round of whole-image generation. When the error is minor, such strategies discard visual content that is already correct; when the error is large, they may carry forward an incorrect layout from the current image and introduce new hallucinations during regeneration.

Figure~\ref{fig:negative_anchor_collateral_damage} shows failure cases observed in our reproduction of prior work~\citep{qin2025unicot}. In the top example, the prompt requires a dining table to the right of an oven, but the initial image places the oven too far to the right and couples it tightly to the scene depth. Image-conditioned repair reuses this layout and cannot accommodate the table farther to the right. In the bottom examples, the same image condition causes collateral errors: the model fails to turn coal into carbon ash, while a requested percussion instrument is fused with the pipa in the original image and loses its proper structure. These cases reveal two complementary risks---an incorrect image can become a negative visual anchor, and even a small edit can alter content outside its target---showing that repair granularity should depend on how much of the current image is worth preserving.

Figure~\ref{fig:unicot_failure} further shows how repeated whole-image regeneration can compound errors. In the top sequence, five correction rounds fail to remove the floating vase while repeatedly rewriting the initially clear sofa and background and accumulating grain noise. In the bottom sequence, the requested phone concept remains recognizable, but the hand and image texture progressively deform. Repair actions should therefore control their disruption scope explicitly: local repair can preserve valid regions and reduce newly introduced artifacts when only a limited area is wrong.

We propose MLIE-CoT, a new framework that converts visual reflection into a sequence of executable actions. It structures visual chain-of-thought as a test-time control process that observes the image, diagnoses errors, selects an action, and performs repair. Unlike textual CoT, the intermediate states of MLIE-CoT are actual images; unlike prompt rewriting alone, every step in MLIE-CoT is grounded in a concrete visual action. MLIE-CoT first uses L0 to accept satisfactory images; otherwise, it routes among L1 local/object-level repair, L2 image-conditioned whole-image editing, and L3 text-only regeneration. These four actions are ordered from least to most disruptive, allowing MLIE-CoT to discard the image condition only when the current image acts as a negative anchor. For limited local errors such as incorrect colors or counts, the framework preserves correct content whenever possible and repairs only the necessary regions.

Allowing an MLLM to choose a repair strategy on its own provides no learned guidance, often requires expensive trial and error, and does not necessarily produce stable decisions. We therefore collect single-action repair results and use their measured action scores as supervision. At deployment time, the router input is limited to the current image, the target prompt, and a standardized fact-based diagnosis with a fixed structure. This learned level router changes the central decision in MLIE-CoT from prompt-based judgment to data-driven repair routing, while using action scores only for training supervision.

We make three main contributions. First, we incorporate local repair into chain-of-thought reasoning for image generation, further unlocking the potential of intermediate images as reusable states. Second, we propose MLIE-CoT, which combines and routes among multi-level repair strategies. Its four fine-grained actions reduce new hallucinations introduced by untargeted prompt rewriting or whole-image regeneration. Third, we train and evaluate a repair-strategy router that learns the mapping from image states to repair granularity; experiments across multiple settings show that the router consistently improves image quality while providing better control over computational cost.

\section{Related Work}
\paragraph{Image Generation and Editing Models.}
Text-to-image generation has progressed from adversarial and autoregressive architectures to diffusion models. AttnGAN uses word-level attention in multi-stage adversarial generation \citep{xu2018attngan}, while DALL-E autoregressively models a unified sequence of text and discrete image tokens \citep{ramesh2021zeroshot}. Imagen combines a large pretrained language encoder with cascaded diffusion \citep{saharia2022photorealistic}, and latent diffusion performs text-conditioned denoising in a compressed space, enabling efficient high-resolution systems such as Stable Diffusion \citep{rombach2022highresolution}. Together, these shifts improve text--image alignment, fidelity, and efficiency, making generation systems increasingly suitable as components of iterative pipelines rather than stand-alone samplers.

Conditioning interfaces have also expanded beyond one-shot generation from text and noise. ControlNet introduces spatial controls such as edges, depth, and pose, while InstructPix2Pix conditions editing on both an image and a natural-language instruction \citep{zhang2023controlnet,brooks2023instructpix2pix}. The task thus extends from describing a new image to continuing or modifying existing visual content. The source image supplies structure and semantics that need not be regenerated. More recent unified models directly support understanding, generation, and editing: BAGEL learns from mixed text, image, and video data and supports free-form editing \citep{deng2025bagel}, while FLUX.2 supports text-to-image, single-reference, and multi-reference editing \citep{blackforestlabs2025flux2}. This shift turns generated images into reusable conditions, providing the model basis for MLIE-CoT's full-image and local-crop repairs.

\paragraph{Visual Chain-of-Thought and Reasoning-Enhanced Generation.}
The earlier PromptCoT uses an adapted textual chain of thought to combine prompt continuations and revisions, producing a text condition that better matches the generator's preferred distribution before synthesis \citep{yao2024promptcot}. ImageGen-CoT constructs reasoning chains for text-to-image in-context learning \citep{liao2025imagegencot}, while "Can We Generate Images with CoT?" studies stepwise verification, potential assessment, and reinforcement \citep{guo2025imagecot}. Uni-CoT unifies macro- and micro-level CoT in an understanding-and-generation model \citep{qin2025unicot}. Although these methods show the value of reasoning, they largely express it through text, embeddings, rewards, or repeated generation. Repeated full-image generation can discard regions that already satisfy the prompt. MLIE-CoT instead treats generated images as visual states, preserving correct regions while assigning the current full image and local crops different roles in later reasoning.

\paragraph{Hierarchical Visual Conditioning and Global--Local Coordination.}
Complex generation can be broken down across semantic levels or spatial scales. Cascaded diffusion conditions later stages on low-resolution outputs to add detail \citep{ho2022cascaded}, while Visual Chain-of-Thought Diffusion Models generate pixels through an intermediate semantic CLIP embedding \citep{voleti2023visualcotdiffusion}. These methods establish intermediate visual states as useful generation conditions. ElasticDiffusion separates global signals for structural consistency from local patch signals for low-level detail \citep{hajiali2024elasticdiffusion}. Similarly, the Describe Anything Model combines global context with a high-resolution focus region for region-level understanding \citep{lian2025describeanything}. The full image provides composition and surrounding context, while crops retain detail that resizing can hide. This pairing is especially useful when an error is limited to one region but its correction must remain consistent with the scene. MLIE-CoT applies this global--local combination to post-generation repair: the current image preserves layout and correct content, while crops focus editing on incorrect regions. Rather than following a fixed coarse-to-fine process, it routes among local repair, whole-image editing, and text-only regeneration according to error scope.

\paragraph{Local Synthesis and Automated Post-Generation Correction.}
Earlier local editing methods relied on explicit regions or layers. With a user-provided mask, Blended Diffusion combines local generation with a noised original during reverse diffusion to preserve unmasked content \citep{avrahami2022blended}; Text2LIVE instead places a generated RGBA editing layer onto the image \citep{bartal2022text2live}. LLM-grounded Diffusion generates foreground objects separately, extracts their masks, and combines their latent features on a global canvas \citep{lian2024llmgrounded}. Self-correcting LLM-controlled Diffusion Models automate this process: a detector and an LLM compare current and target layouts, translate differences into object-level operations, and use segmentation and diffusion to repair blending errors \citep{wu2024selfcorrecting}. ZONE also finds an instructed region, extracts a local layer, and puts it back with edge smoothing \citep{li2024zone}. These methods close the loop from detection to correction, but still rely on fixed local operations. MLIE-CoT also decides whether an observed failure is best handled locally or requires a broader repair. These approaches show that restricting edits to incorrect regions helps preserve correct content, motivating MLIE-CoT's local repair level.

\section{The MLIE-CoT Method}

\begin{figure*}[t]
  \centering
  \includegraphics[width=0.98\textwidth]{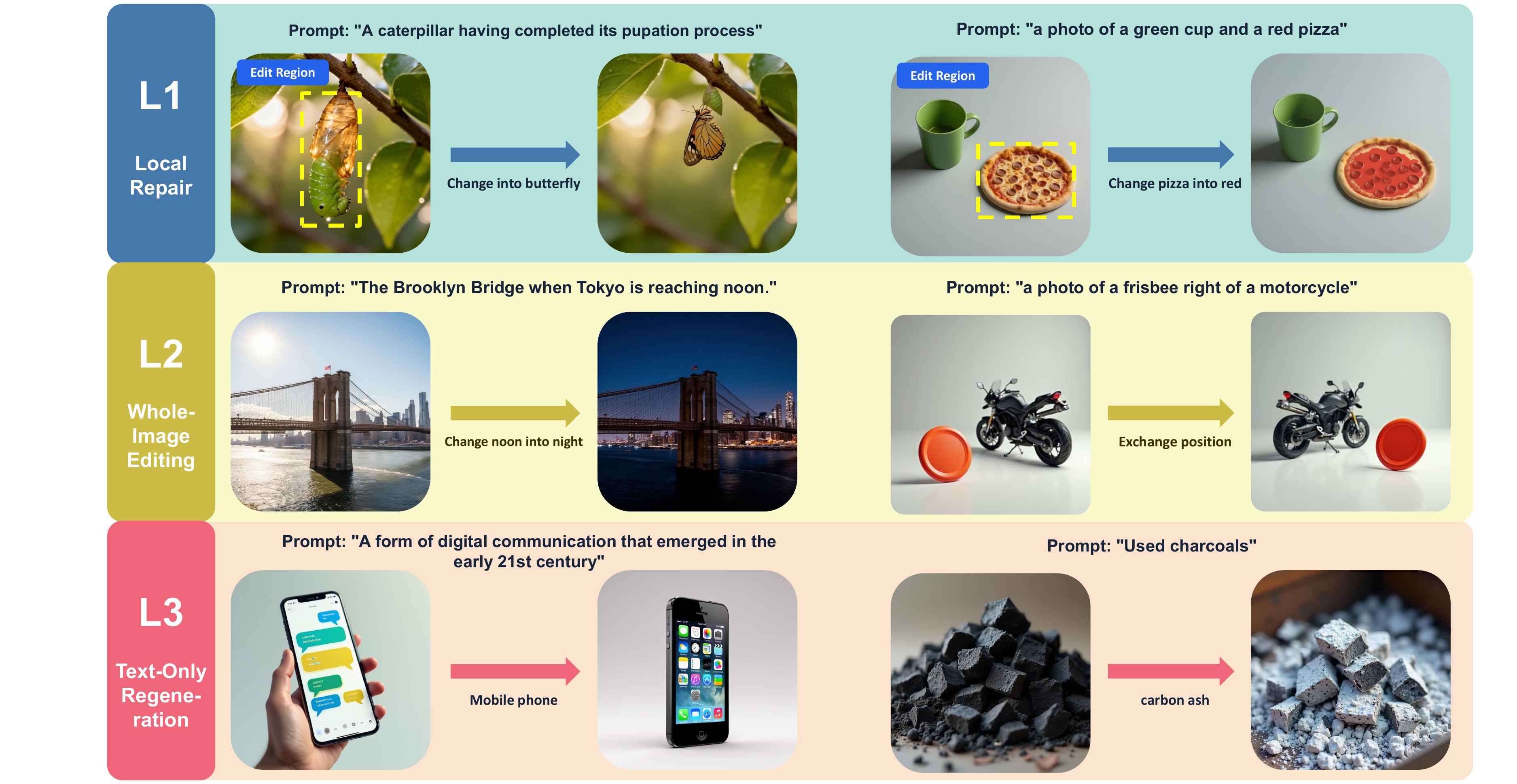}
  \caption{Representative results for the three routed repair levels.}
  \label{fig:level_examples}
\end{figure*}

\begin{figure*}[t]
  \centering
  \includegraphics[width=0.95\textwidth]{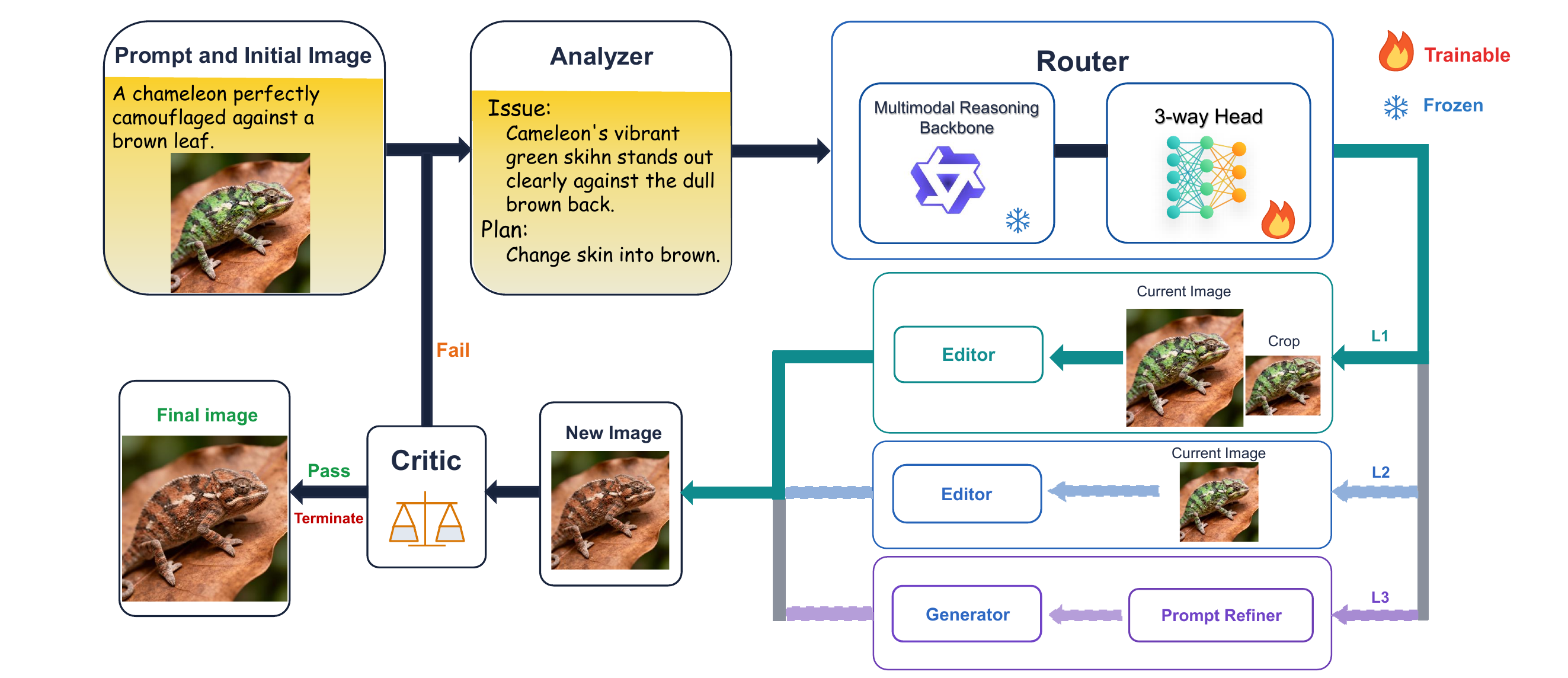}
  \caption{Overview of the MLIE-CoT framework. The analyzer identifies mismatches between the prompt and current image. A frozen multimodal reasoning backbone feeds its representation to a trainable three-way prediction head, which selects L1 local editing with cropped regions, L2 whole-image editing conditioned on the current image, or L3 regeneration by the prompt refiner and generator. The critic examines the new image: a pass returns the final image and terminates, whereas a failure sends the result back for another diagnosis and repair round.}
  \label{fig:pipeline}
\end{figure*}

\subsection{Problem Formulation and Framework Overview}

Multi-level Image Editing Chain-of-Thought (MLIE-CoT) organizes post-generation inspection and repair as a sequence of visual-state updates. Given a target prompt $p$, the system first applies a target-reasoning and prompt-optimization procedure $Q$ to obtain a prompt $p_g$ better suited to the generation model, and then samples an initial image $x_0$ from the text-to-image model $G_0$. We define the state at iteration $t$ by the target, the current image, and the history:
\begin{equation}
  p_g=Q(p), \qquad x_0\sim G_0(p_g), \qquad s_t=(p,x_t,h_t),
  \label{eq:initial_generation}
\end{equation}
where $h_t$ records previously executed actions, review outcomes, and the iteration count. At each iteration, the diagnostic model $F_{\mathrm{ref}}$ examines the current image $x_t$ against the target $p$ and produces a fixed-schema factual diagnosis $d_t$. The diagnosis contains a constraint-by-constraint assessment, visible evidence, observed and expected values, and an acceptance flag $q_t$:
\begin{equation}
  \begin{aligned}
    (d_t,q_t) &= F_{\mathrm{ref}}(p,x_t).\quad
  \end{aligned}
  \label{eq:reflection}
\end{equation}
When $q_t=1$, the system accepts the current image and terminates. If $q_t=0$ and an action budget remains, the router $R_\theta$ predicts a repair level $\hat l_t$. The planner $\mathcal{P}$ then produces a level-appropriate repair plan $e_t$, the editor $A_{l_t}$ generates a repaired image $\widetilde{x}_{t+1}$, and the reviewer $V$ selects the next state from the original and repaired images:
\begin{equation}
  \begin{aligned}
    \hat l_t &= R_\theta(x_t,p,d_t),
      &&\hat l_t\in\{1,2,3\},\\
    (l_t,e_t) &= \mathcal{P}(\hat l_t;p,x_t,d_t),\\
    \widetilde{x}_{t+1} &= A_{l_t}(p,x_t,e_t),\\
    (v_t,x_{t+1}) &= V(p,d_t,e_t,x_t,\widetilde{x}_{t+1}),
      &&x_{t+1}\in\{x_t,\widetilde{x}_{t+1}\},\\
    h_{t+1} &= h_t\oplus r_t.
  \end{aligned}
  \label{eq:visual_cot_transition}
\end{equation}
The review outcome is stored in $v_t$, while the diagnosis, route, plan, repaired image, and final selection from the current iteration are combined in the record $r_t$.

\subsection{Hierarchical Repair and Repair Acceptance}

Different errors require changes at different scopes. Local issues involving color, attributes, or a small number of objects generally favor targeted repair. Background, spatial-relation, and layout errors call for whole-image editing. When the main concept, multi-object composition, or overall arrangement is incorrect, the current image itself may repeatedly mislead the editor, making text-only regeneration more appropriate. Because no single repair path is best in all cases, MLIE-CoT defines three repair levels with increasingly broad scopes. These levels describe how much of the current image is preserved rather than the difficulty of the error, allowing the router to adjust intervention strength while reusing the same diagnostic representation as the visual state evolves. Table~\ref{tab:levels} summarizes their intended use, and Figure~\ref{fig:level_examples} presents typical examples.

\begin{table}[t]
  \centering
  \small
  \setlength{\tabcolsep}{3pt}
  \begin{tabular}{@{}p{0.09\columnwidth}>{\raggedright\arraybackslash}p{0.26\columnwidth}p{0.57\columnwidth}@{}}
    \toprule
    Level & Action & Typical scenario \\
    \midrule
    L0 & Accept current image & The current image satisfies the constraints in the target prompt. \\
    L1 & Local repair & Errors in color, attributes, counts, or a few objects that affect only a small number of regions. \\
    L2 & Whole-image editing & Spatial relations, background, or layout require adjustment, but the current image remains worth preserving. \\
    L3 & Text-conditioned regeneration & The main concept or overall composition is incorrect, or multiple connected errors are difficult to edit one by one. \\
    \bottomrule
  \end{tabular}
  \caption{Hierarchical actions in MLIE-CoT.}
  \label{tab:levels}
\end{table}

The overall procedure is illustrated in Figure~\ref{fig:pipeline}. The planner performs a slightly different role at each repair level. An L1 plan specifies $m_t\in\{1,2,3\}$ bounding boxes and crops the corresponding localization images $c_{t,1:m_t}$ from the current image. The local repair model $E_{\mathrm{loc}}$ receives the ordered input $[x_t,c_{t,1:m_t}]$, with the full image placed first. L2 uses the whole-image editor $E_{\mathrm{glob}}$, which takes the current image $x_t$ and the editing plan $e_t$. L3 passes a new text condition $e_t$ to the initial text-to-image model $G_0$ without providing the current image. Formally,
\begin{equation}
  \widetilde{x}_{t+1}=
  \begin{cases}
    E_{\mathrm{loc}}([x_t,c_{t,1:m_t}],e_t),
      & l_t=1,\;1\leq m_t\leq3,\\
    E_{\mathrm{glob}}(x_t,e_t), & l_t=2,\\
    G_0(e_t), & l_t=3.
  \end{cases}
  \label{eq:hierarchical_repair}
\end{equation}

After repair, the reviewer compares the images before and after the action and retains the one that better satisfies the prompt, preventing an unsuitable repair from degrading image quality.

\subsection{Utility-Supervised Repair-Level Routing}

Trying every repair level at each iteration would require three expensive generation calls. We instead learn a router that predicts the level with the highest expected one-step utility from the current image $x$, target prompt $p$, and factual diagnosis $d$, so only one repair is generated at test time.

The router uses a frozen Qwen3.5-9B multimodal encoder~\citep{qwen2026qwen359b} and a trainable two-layer classification head $h_\theta$. Its decision is
\begin{equation}
  \begin{aligned}
    z&=f_{\mathrm{enc}}(x,p,d),\\
    \pi_\theta&=\operatorname{softmax}(h_\theta(z)),\\
    \hat l&=\arg\max_{l\in\{1,2,3\}}\pi_\theta(l).
  \end{aligned}
  \label{eq:router_summary}
\end{equation}
This compact design retains the multimodal backbone's understanding of the prompt and image while learning only the relative preference among L1--L3.

\paragraph{Constructing Repair-Level Supervision.}
We construct target prompts with atom compositional words and phrases. The prompts cover common objects and co-occurrences, counts, colors, spatial positions, and attribute binding. Relying only on naturally occurring generation errors would make collection inefficient, so we create controlled errors by replacing selected components of each construction prompt. We further include joint relation and attribute constraints, scene and composition changes, multiple simultaneous constraints, and full semantic replacements. For target prompt $p_n$, a rewrite $a_n$ from the predefined set $\mathcal{A}$ produces the initial-generation prompt $p_n^{\mathrm{init}}$:
\begin{equation}
  \begin{aligned}
    p_n^{\mathrm{init}} &= \mathcal{M}_{a_n}(p_n),\\
    x_n &\sim G_0(p_n^{\mathrm{init}}),\\
    d_n &= F_{\mathrm{ref}}(p_n,x_n).
  \end{aligned}
  \label{eq:router_state_collection}
\end{equation}
The perturbed prompt is used only to generate $x_n$, whereas the unmodified $p_n$ remains the repair target. Images that already satisfy the target are assigned to L0 and excluded from three-way routing; only states $S_n=(x_n,p_n,d_n)$ with visible errors are retained. We split the data by target-prompt family so that rewrites and repeated samples of the same target cannot cross the training, validation, and test sets.

For each retained state, L1--L3 use the same planners and prompt templates as the online method. The levels share the initial image, target, and diagnosis but use level-specific action prompts $e_{n,l}=P_l(p_n,x_n,d_n)$. Each level is executed independently $K$ times, with its plan fixed and the random seed shared across levels for each repeat:
\begin{equation}
  \begin{aligned}
    \widetilde{x}_{n,l,k}
      &=A_l(p_n,x_n,e_{n,l};\xi_{n,k}),\\
    &\quad l\in\{1,2,3\},\quad k=1,\ldots,K.
  \end{aligned}
  \label{eq:counterfactual_rollouts}
\end{equation}
The construction record and factual diagnosis define verifiable yes/no questions for target satisfaction, diagnosed-error correction, and preservation of existing correct content. Qwen answers these questions and flags artifacts or newly introduced errors; the program then deterministically computes the score proportions and the utility of each repair:
\begin{equation}
  \begin{aligned}
    u_{n,l,k}={}&
      \alpha_{\mathrm{gain}}\!\left(q^{\mathrm{cand}}_{n,l,k}
      -q^{\mathrm{orig}}_{n,l,k}\right)
      +\alpha_{\mathrm{fix}}q^{\mathrm{fix}}_{n,l,k}
      +\alpha_{\mathrm{pres}}q^{\mathrm{pres}}_{n,l,k}\\
      &-\alpha_{\mathrm{new}}\nu^{\mathrm{new}}_{n,l,k}
      -\alpha_{\mathrm{art}}a_{n,l,k}
      -\alpha_{\mathrm{cost}}c_{n,l,k},\\
    \bar u_{n,l}={}&\frac{1}{K}\sum_{k=1}^{K}u_{n,l,k}.
  \end{aligned}
  \label{eq:rollout_utility}
\end{equation}
Here, $c_{n,l,k}$ is the number of executor calls, and the six $\alpha$
coefficients are nonnegative utility-weight hyperparameters. A
state is retained only when every level yields exactly $K$ verifiable results,
ensuring that each target is computed from a complete and symmetric set of
$3K$ repair outcomes.

The average utilities are converted into soft level targets. The router is trained with a composite objective that combines soft-target classification, utility alignment, and pairwise level preference:
\begin{equation}
  \mathcal{L}_{\mathrm{route}}
  =\lambda_{\mathrm{soft}}\mathcal{L}_{\mathrm{soft}}
  +\lambda_{\mathrm{util}}\mathcal{L}_{\mathrm{util}}
  +\lambda_{\mathrm{pair}}\mathcal{L}_{\mathrm{pair}}
  +\lambda_{\mathrm{eff}}\mathcal{L}_{\mathrm{eff}}.
  \label{eq:router_loss_summary}
\end{equation}
Here, $\mathcal{L}_{\mathrm{eff}}$ resolves exact quality ties in favor of the
lower-cost action. The full routing-head parameterization, question-count
equations, loss definitions, coefficient values, and implementation hyperparameters are provided
in the supplementary material.

\section{Experiments}

\subsection{Training Setup}
We construct 1K prompts and sample each repair level five times, yielding a routing-training dataset of 14.8K images. We freeze Qwen3.5-9B~\citep{qwen2026qwen359b} and train only the two-layer routing head with the composite objective in Eq.~\eqref{eq:router_loss_summary}. Full optimization and architecture settings are reported in the supplementary material.

\subsection{Experimental Setup and Configuration}

\paragraph{Evaluation Benchmarks}
We mainly evaluate the improvements of MLIE-CoT in text-to-image semantic alignment and reasoning-intensive generation using GenEval~\citep{ghosh2023geneval} and WISE~\citep{niu2025wise}. GenEval contains 553 object-centered prompts covering single objects, two objects, counting, colors, spatial positions, and color-attribute binding. Its official evaluator uses object detection and instance segmentation to determine whether all constraints are satisfied together, and Overall is the average of the six category accuracies. WISE contains 1,000 prompts covering cultural knowledge, time, space, biology, physics, and chemistry. Unlike compositional prompts that directly name the target, WISE often specifies it indirectly through knowledge clues, space-time relations, or scientific conditions, testing whether a model can first recover the hidden knowledge and then correctly show the inferred conclusion in the image.

We evaluate BAGEL-7B-MoT~\citep{deng2025bagel} and FLUX.2-dev~\citep{blackforestlabs2025flux2} as generation or editing backbones. The following experiments are all carried out on the official prompts and evaluators.

\subsection{Main Results}

\subsubsection{GenEval}

\begin{table*}[t]
\centering
\small
\setlength{\tabcolsep}{2pt}
\begin{tabular*}{\textwidth}{@{\extracolsep{\fill}}llccccccc@{}}
\toprule
Group & Model/Method & Single Obj. & Two Obj. & Count. & Color & Position & Attr. Bind. & Overall \\
\midrule
Base model & BAGEL & 0.99 & 0.92 & 0.78 & 0.87 & 0.56 & 0.66 & 0.80 \\
 & FLUX.2-dev & 0.98 & 0.91 & 0.69 & 0.95 & 0.80 & 0.73 & 0.84 \\

\midrule
Related method & Uni-CoT~\citep{qin2025unicot} & 0.99 & \textbf{0.96} & \textbf{0.84} & 0.92 & 0.57 & 0.71 & 0.83 \\
 & Image Generation CoT (best PARM)~\citep{guo2025imagecot} & 0.99 & 0.86 & 0.67 & 0.84 & 0.66 & 0.64 & 0.77 \\
\midrule
Ours & MLIE-CoT + BAGEL  &0.97 & 0.94 & 0.74 & \textbf{0.96} & 0.80 & 0.75 & 0.86\\
 & MLIE-CoT + FLUX.2-dev  & \textbf{1.00} & 0.93 & 0.79 & 0.93 & \textbf{0.86} & \textbf{0.80} & \textbf{0.88} \\
\bottomrule
\end{tabular*}
\caption{GenEval test results.}
\label{tab:geneval-main}
\end{table*}

Table~\ref{tab:geneval-main} shows that MLIE-CoT improves overall compositional generation with both base models, raising Overall from 0.80 to 0.86 for BAGEL and from 0.84 to 0.88 for FLUX.2-dev. The gains are especially pronounced for spatial position and attribute binding.

\subsubsection{WISE}

\begin{table*}[t]
\centering
\small
\setlength{\tabcolsep}{2pt}
\begin{tabular*}{\textwidth}{@{\extracolsep{\fill}}llccccccc@{}}
\toprule
Group & Model/Method & Culture & Time & Space & Biology & Physics & Chemistry & Overall \\
\midrule
Base model & BAGEL & 0.76 & 0.69 & 0.75&  0.60 & 0.66 & 0.63 & 0.70 \\
 & FLUX.2-dev & 0.80 & 0.67 & 0.83 & 0.63 & 0.64 & 0.63 & 0.73 \\
\midrule
Local baseline & ImageGen without CoT (SEED-X) & 0.60 & 0.60 & 0.56 & 0.48 & 0.55 & 0.34 & 0.55 \\
 & ImageGen-CoT (one CoT chain)~\citep{liao2025imagegencot} & 0.65 & 0.61 & 0.65 & 0.49 & 0.53 & 0.34 & 0.58 \\
 & Uni-CoT~\citep{qin2025unicot} & 0.79 & 0.69 & 0.83 & \textbf{0.67} & 0.67 & 0.70 & 0.74 \\
\midrule
Ours & MLIE-CoT + BAGEL & \textbf{0.83} & 0.69 & 0.86 & 0.63 & 0.68 & 0.71 & 0.76  \\
 & MLIE-CoT + FLUX.2-dev & 0.82 & \textbf{0.75} & \textbf{0.88} & \textbf{0.67} & \textbf{0.70} & \textbf{0.77} & \textbf{0.78} \\
\bottomrule
\end{tabular*}
\caption{WISE test results.}
\label{tab:wise-main}
\end{table*}

Table~\ref{tab:wise-main} extends this finding to tasks that require recovering implicit knowledge before expressing it visually. MLIE-CoT raises Overall from 0.70 to 0.76 with BAGEL and from 0.73 to 0.78 with FLUX.2-dev. The comparable gains on both backbones, together with improvements across several categories, show that the mechanism generalizes across base models and knowledge categories.

\subsection{Ablation Studies}

\subsubsection{Independent Gains and the Added Benefit of Combining the Three Repair Actions}
To test whether combining multiple repair strategies is necessary, we compare against fixed repair policies. The generation backbone is FLUX.2-dev, and the vision-language backbone shared by the analyzer and router is a frozen Qwen3.5-9B. All four policies reuse exactly the same set of 553 initial images at a resolution of $1024\times1024$ on a per-sample basis rather than resampling them independently; differences in the table therefore arise only from the subsequent repair policy. L1-to-L3-only denotes a setting in which the model iteratively refines images with a predefined repair strategy. Self-route denotes a model-driven inference routing strategy that dynamically selects the repair process based on the coarse-grained image repair taxonomy introduced in Table 1.

\begin{table}[t]
\centering
\small
\setlength{\tabcolsep}{0.5pt}
\begin{tabular*}{\columnwidth}{@{\extracolsep{\fill}}lccccccc@{}}
\toprule
Policy & 1-Obj. & 2-Obj. & Cnt. & Col. & Pos. & Bind. & Avg. \\
\midrule
Init.   & 0.98 & 0.91 & 0.69 & \textbf{0.95} & 0.80 & 0.73 & 0.84 \\
L1-only & \textbf{1.00} & 0.91 & 0.78 & 0.92 & 0.76 & 0.78 & 0.86 \\
L2-only & \textbf{1.00} & 0.91 & 0.78 & 0.93 & 0.73 & 0.78 & 0.85 \\
L3-only & \textbf{1.00} & 0.91 & \textbf{0.79} & 0.92 & 0.73 & 0.78 & 0.85 \\
self-route &\textbf{1.00} & \textbf{0.94} & 0.71 & \textbf{0.95} & 0.81 & 0.74 & 0.85   \\
MLIE & \textbf{1.00} & 0.93 & \textbf{0.79} & 0.93 & \textbf{0.86} & \textbf{0.80} & \textbf{0.88} \\
\bottomrule
\end{tabular*}
\caption{Fixed-level and router ablations on the same set of initial images.}
\label{tab:fixed-level-ablation}
\end{table}

With the same initial images strictly shared across policies, Table~\ref{tab:fixed-level-ablation} shows that all three fixed actions improve the results but suit different scopes: L1 preserves content while making local corrections, L2 coordinates global attributes and layout, and L3 resets samples that cannot be repaired reliably. Router further raises accuracy to 88\%. Its advantage therefore comes from deciding how much information from the current image to preserve according to error severity; this error heterogeneity is the fundamental reason multi-level routing outperforms fixed repair.

\subsection{Effect and Complexity Analysis of the Number of Chain-of-Thought Steps}

\begin{figure}[t]
  \centering
  \includegraphics[width=\columnwidth]{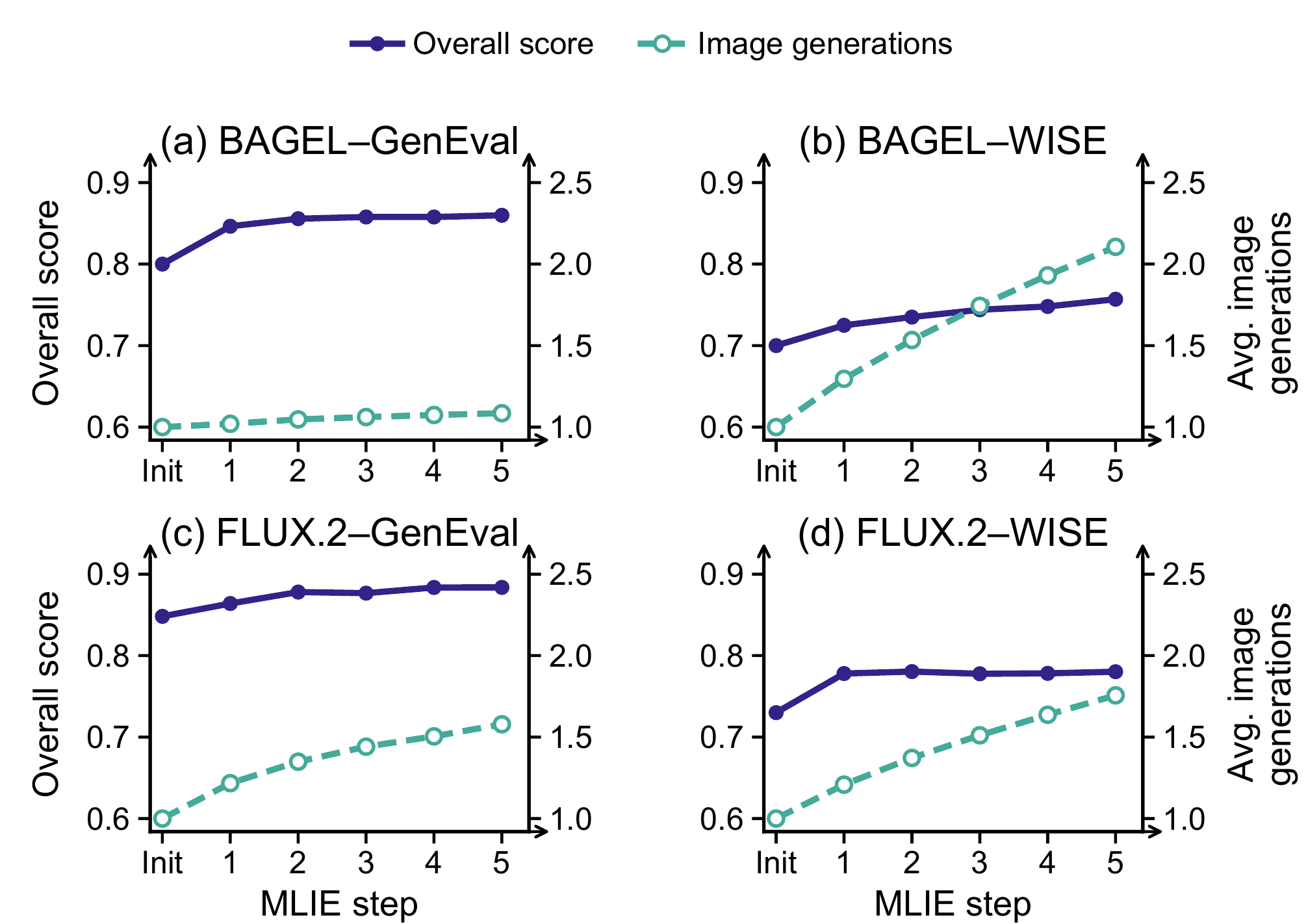}
  \caption{Effect of the number of MLIE-CoT steps on generation quality and cost for BAGEL and FLUX.2 on GenEval and WISE.}
  \label{fig:step_effect_complexity}
\end{figure}

Chain-of-thought methods typically use additional reasoning and incur extra computation in exchange for better generation quality, and the length of chain effect the cost most. We design an ablation experiment on Chain-of-Thought Steps. We use generation count to represent the added cost as analysis and routing are negligible relative to generation, while wall-clock time is hardware- and implementation-dependent. Figure~\ref{fig:step_effect_complexity} shows that larger step budgets generally improve or maintain performance, but the gains have a clear ceiling. Increasing the repair budget gives difficult samples more opportunities to correct remaining errors, but it does not guarantee that performance will continue to rise as more steps are used. The resulting benefit depends jointly on the reasoning demands of the task and the correctable error space left by the base model. The main value of a larger step budget is therefore not unlimited score gains, but broader coverage of difficult samples that may require additional repair attempts.

Quality and cost therefore do not scale linearly with the number of repair steps. A small number of steps provides a cost-effective default, whereas raising the upper limit allows the system to attempt harder tasks at the price of additional computation. Because the review-and-accept mechanism preserves an improved result and rejects unsuccessful revisions, we do not observe obvious noise accumulation or image-quality degradation over multiple repair steps. Nevertheless, inherent model noise and capacity limits mean that additional steps do not necessarily yield higher quality and may merely add cost. Our results therefore suggest that one to three repair steps offer the most efficient operating range.

For context, we also compare the cost with that of other methods for optimizing image generation. In our reproduction using the default settings, Uni-CoT generates an average of 1.107 images per prompt on GenEval and 1.449 images per prompt on WISE. These results indicate that the additional generation cost introduced by our method remains broadly comparable to that of existing chain-of-thought methods, rather than requiring a substantially larger number of image-generation calls.

\FloatBarrier

\section{Conclusion}

MLIE-CoT combines local repair, whole-image editing, and text-only regeneration to reduce hallucination accumulation and local-detail errors. A strict same-initial-image GenEval ablation shows that router outperforms any fixed action and self-route. The framework is model-agnostic: native multimodal models can serve directly as action tools, while models that support reference-image inputs can serve as image-generation backbones. In the future, targeted optimization of repair prompts is a promising direction. Current models also generate at fixed, limited resolutions. Image-generation CoT may help extend them to larger images, particularly for complex tasks that cannot be completed reliably in a single generation.

\clearpage
\bibliography{references}

\clearpage
\appendix
\setcounter{secnumdepth}{1}
\setcounter{equation}{0}
\setcounter{table}{0}
\setcounter{figure}{0}
\renewcommand{\theequation}{S\arabic{equation}}
\renewcommand{\thetable}{S\arabic{table}}
\renewcommand{\thefigure}{S\arabic{figure}}
\twocolumn[\begin{center}\Large\bfseries Supplementary Material\end{center}\vspace{1em}]
\section{Router Training Details}
\label{sec:router_training_details}

\subsection{Architecture and Inference}

The frozen Qwen3.5-9B backbone~\citep{qwen2026qwen359b} $f_{\mathrm{enc}}$ encodes the current RGB image, target prompt, and factual diagnosis. With left padding, $z$ is the $D$-dimensional last-layer state of the final valid token. The classification head applies layer normalization, a linear layer, GELU, dropout, and a second linear layer:
\begin{equation}
  \begin{aligned}
    \boldsymbol{\phi} &= \operatorname{Dropout}_{\rho}\!\left(
      \operatorname{GELU}\!\left(W_1\operatorname{LN}(z)+b_1\right)
      \right),\\
    \mathbf{o} &= W_2\boldsymbol{\phi}+b_2,\\
    \pi_\theta(l\mid x,p,d) &= \operatorname{softmax}(\mathbf{o})_l,\\
    \hat l &= \arg\max_{l\in\{1,2,3\}}\pi_\theta(l\mid x,p,d).
  \end{aligned}
  \label{eq:route_head}
\end{equation}
The deployed router concatenates the 4,096-dimensional Qwen state with 11
train-normalized, deterministic diagnosis statistics: issue counts for the
seven known constraint types and an ``other'' type, total issue count,
preservation-constraint count, and mean issue confidence.  Thus, the head
input dimension is 4,107.  The head parameters are
$\theta=\{W_1,b_1,W_2,b_2\}$, its hidden dimension is $128$, and
$\rho=0.30$. The three logits correspond to L1--L3; only the head is updated
during training.

\subsection{Detailed Rollout Scoring}

Let $b^{\mathrm{orig}}$, $b^{\mathrm{cand}}$, $b^{\mathrm{fix}}$, and $b^{\mathrm{pres}}$ denote binary answers for the original target, repaired target, diagnosed-error, and preservation questions. Their pass rates are
\begin{equation}
  \begin{aligned}
    q^{\mathrm{orig}}_{n,l,k}
      &=\frac{1}{|\mathcal{Q}^{\mathrm{tar}}_n|}
        \sum_{i\in\mathcal{Q}^{\mathrm{tar}}_n}
        b^{\mathrm{orig}}_{n,l,k,i},\\
    q^{\mathrm{cand}}_{n,l,k}
      &=\frac{1}{|\mathcal{Q}^{\mathrm{tar}}_n|}
        \sum_{i\in\mathcal{Q}^{\mathrm{tar}}_n}
        b^{\mathrm{cand}}_{n,l,k,i},\\
    q^{\mathrm{fix}}_{n,l,k}
      &=\frac{1}{|\mathcal{Q}^{\mathrm{fix}}_n|}
        \sum_{i\in\mathcal{Q}^{\mathrm{fix}}_n}
        b^{\mathrm{fix}}_{n,l,k,i},\\
    q^{\mathrm{pres}}_{n,l,k}
      &=\frac{1}{|\mathcal{Q}^{\mathrm{pres}}_n|}
        \sum_{i\in\mathcal{Q}^{\mathrm{pres}}_n}
        b^{\mathrm{pres}}_{n,l,k,i}.
  \end{aligned}
  \label{eq:question_count_scores}
\end{equation}
We set $q^{\mathrm{pres}}_{n,l,k}=1$ when no explicit preservation condition exists. Qwen also provides binary artifact and new-error flags $a_{n,l,k},\nu^{\mathrm{new}}_{n,l,k}$; the program combines all quantities through the composite rollout utility defined in the main paper. Qwen supplies only verifiable binary answers, while the program deterministically computes scores and labels.
For the utility in Eq.~(8) of the main paper, the coefficient hyperparameters
are
\[
  (\alpha_{\mathrm{gain}},\alpha_{\mathrm{fix}},
  \alpha_{\mathrm{pres}},\alpha_{\mathrm{new}},
  \alpha_{\mathrm{art}},\alpha_{\mathrm{cost}})
  =(3,2,1,3,1,0.15).
\]

\subsection{Composite Utility Objective and Training Loss}

We convert the average utilities of the three levels into soft labels using temperature $\tau_s=0.5$:
\begin{equation}
  y_{n,l}=\frac{\exp(\bar u_{n,l}/\tau_s)}
  {\sum_{j=1}^{3}\exp(\bar u_{n,j}/\tau_s)}.
  \label{eq:soft_route_target}
\end{equation}
To reduce level imbalance, we sum each level's soft-label mass over $\mathcal{D}_{\mathrm{tr}}$ and define
\begin{equation}
  m_l=\sum_{n\in\mathcal{D}_{\mathrm{tr}}}y_{n,l}, \qquad
  w_l=\operatorname{clip}\!\left(
    \frac{|\mathcal{D}_{\mathrm{tr}}|}{3m_l},
    \frac{1}{w_{\max}},w_{\max}\right),
  \label{eq:route_class_weight}
\end{equation}
Here, $\operatorname{clip}(z,a,b)=\min\{\max\{z,a\},b\}$ and $w_{\max}=4$. Validation and test distributions are unchanged.

For each state in a batch of size $N$, logits and utilities are centered across levels:
\begin{equation}
  \widetilde{o}_{n,l}=o_{n,l}-\frac{1}{3}\sum_{j=1}^{3}o_{n,j}, \qquad
  \widetilde{u}_{n,l}=\bar u_{n,l}-\frac{1}{3}\sum_{j=1}^{3}\bar u_{n,j}.
\end{equation}
Utility differences define soft pairwise targets:
\begin{equation}
  \gamma_{n,ij}=\operatorname{sigmoid}\!\left(
    \frac{\bar u_{n,i}-\bar u_{n,j}}{\tau_p}\right).
\end{equation}
With $i<j$, $\tau_u=\tau_p=0.5$. The loss combines class-weighted soft-label cross-entropy, centered-utility alignment, and pairwise binary cross-entropy:
\begin{equation}
  \begin{aligned}
    \mathcal{L}_{\mathrm{soft}}
      &=-\frac{1}{N}\sum_{n=1}^{N}\sum_{l=1}^{3}
        w_l y_{n,l}\log\pi_\theta(l\mid x_n,p_n,d_n),\\
    \mathcal{L}_{\mathrm{util}}
      &=\frac{1}{3N}\sum_{n=1}^{N}\sum_{l=1}^{3}
        \operatorname{SmoothL1}\!\left(
          \widetilde{o}_{n,l},\frac{\widetilde{u}_{n,l}}{\tau_u}\right),\\
    \mathcal{L}_{\mathrm{pair}}
      &=\frac{1}{3N}\sum_{n=1}^{N}\sum_{1\leq i<j\leq3}
        \ell_{\mathrm{BCE}}(o_{n,i}-o_{n,j},\gamma_{n,ij}),\\
    \mathcal{L}_{\mathrm{route}}
      &=0.50\mathcal{L}_{\mathrm{soft}}
        +0.10\mathcal{L}_{\mathrm{util}}
        +0.20\mathcal{L}_{\mathrm{pair}}
        +0.001\mathcal{L}_{\mathrm{eff}}.
  \end{aligned}
  \label{eq:route_loss}
\end{equation}
The pairwise term covers all three unordered level pairs.
The coefficient values used in Eq.~(9) of the main paper are
$(\lambda_{\mathrm{soft}},\lambda_{\mathrm{util}},
\lambda_{\mathrm{pair}},\lambda_{\mathrm{eff}})
=(0.50,0.10,0.20,0.001)$.
$\mathcal{L}_{\mathrm{eff}}$ breaks exact quality ties in favor of the
lower-cost action, using measured L1--L3 costs of
$(204.5771,154.4044,151.5316)$ seconds.  Non-tied quality comparisons are
unaffected by this term.  The policy temperature is $1.0$, the quality-tie
tolerance is zero, and the expected-regret and uncertain-safe-prior weights
are both zero.

\subsection{Implementation and Optimization}

The 1K prompts, five samples per level, and three levels yield 15K images.
Qwen3.5-9B is frozen and loaded in 8-bit precision; its final non-padding
token provides a 4,096-dimensional feature. We train the two-layer head with
AdamW~\citep{loshchilov2019decoupled}: learning rate $10^{-4}$, weight decay
$10^{-2}$, batch size $256$, at most $300$ epochs, early-stopping patience
$40$, minimum improvement $10^{-6}$, and gradient-norm cap $5.0$.  The
training seed is 20260719; it initializes Python, PyTorch, and all CUDA
generators, while epoch $e$ uses seed $20260719+e$ for the training
permutation.  Checkpoint selection minimizes validation quality regret, then
measured action cost, then validation loss; the selected checkpoint is from
epoch 110.  The test split is not used for checkpoint selection.

\section{Reproducibility and Execution Details}
\label{sec:reproducibility_details}

\subsection{Computing Infrastructure}

Experiments were executed on a Linux node with four NVIDIA
A100-SXM4-80GB GPUs (80\,GiB per GPU), two Intel Xeon Platinum 8362 CPUs
(32 cores per socket and 128 hardware threads in total), and 935\,GiB of
visible system memory.  The operating system is Ubuntu 22.04.5 LTS with
Linux kernel 5.4.250.  The recorded NVIDIA driver is 535.129.03;
\texttt{nvidia-smi} reports CUDA 12.2 driver capability, whereas the PyTorch
wheel was built with CUDA 12.8.  The software environment uses Python
3.10.12, PyTorch 2.10.0 (cu128), cuDNN 9.10.2, torchvision 0.25.0,
Transformers 5.5.3, Diffusers 0.37.1, Accelerate 1.13.0,
bitsandbytes 0.49.2, Pillow 12.2.0, NumPy 2.2.6, Safetensors 0.7.0,
and Hugging Face Hub 1.11.0.

\subsection{Inference Configuration}

The analyzer and router use Qwen3.5-9B as their backbone. The model offers
strong multimodal understanding and has a parameter count comparable to
those of FLUX.2-dev and BAGEL, which helps avoid scaling-law confounds when
it is used as a component of the reasoning chain. We use immutable revision
\texttt{c2022362\allowbreak35762e1c\allowbreak871ad0cc\allowbreak
b60c8ee5\allowbreak ba337b9a}, loaded in BF16 with
8-bit weights and no CPU offload.  Analyzer inputs have maximum long side
1,536 and maximum 1,536 new tokens; review and comparison images have maximum
long side 1,024.  Normal decoding uses \texttt{do\_sample=false}; the
deterministic malformed-output retry additionally uses repetition penalty
1.05 and no-repeat n-gram size 8.

FLUX.2-dev repair calls use immutable revision
\texttt{d72c4754\allowbreak e67efcac\allowbreak a88832fc\allowbreak
18ea6945\allowbreak47a28bd2}, BF16, a
$1024\times1024$ output, 28 inference steps, guidance scale 3.5, maximum
sequence length 512, and one candidate per action.  L1 supplies the editable
full image followed by one to three planner-ordered localization crops, uses
crop padding ratio 1.35 and minimum crop span 64 pixels, and resizes every
crop to the generation resolution.  L2 supplies only the editable full
image.  L3 generates from rewritten text without an image reference.  L1--L3
make one semantic executor call and do not generate extra semantic retry
candidates; infrastructure failures retry the exact same request.  The
maximum repair budget is five generated actions per prompt, and candidate
review does not consume this budget.

\subsection{Randomness and Number of Runs}

The base seed is 42.  For zero-based prompt index
$i$ and one-based repair round $r$, the initial-generation seed is $42+i$,
the L2 whole-image edit seed is $42+100i+r+30000$, and the L3 regeneration
seed is $42+100i+r$.  L1 uses a stable 31-bit seed obtained by taking the
first eight bytes of SHA-256 over canonical JSON containing the action-schema
version, base seed 42, sample identifier, repair round, and ordered edit
specifications, then reducing modulo $2^{31}-1$.  Every FLUX.2 request
constructs an isolated device-local
\texttt{torch.Generator(...).manual\_seed(seed)}.  Thus, scheduling a prompt
on a different one of the four GPUs does not change its assigned seed.

Each main-table entry for our method is computed from one complete generation
trajectory over all official prompts for the corresponding benchmark and configuration:
553 prompts for GenEval and 1,000 unique prompt IDs for WISE.  The
fixed-level ablation reuses the same initial image for every policy on a
per-prompt basis.  Router-data collection is different: each retained state
uses five paired stochastic rollouts per repair level, with the seed shared
across L1--L3 within a repeat.  The separate WISE judge-stability audit uses
four evaluator seeds (20260725--20260728) on identical frozen images; these
are evaluator repetitions and are not additional generation trajectories.

\subsection{Evaluation Aggregation}

GenEval reports the official mean for each of its six task categories and
defines Overall as the unweighted mean of those six category means.  WISE
uses one valid binary decision for each unique prompt ID 1--1000; Overall is
the mean of those 1,000 decisions, and category scores are means over the
corresponding official ID subsets.  We do not label an incomplete or
duplicate-ID WISE evaluation as an official Overall score.  These benchmarks
were selected because GenEval measures explicit compositional constraints,
whereas WISE measures whether implicit cultural, spatiotemporal, and
scientific knowledge is correctly recovered and depicted.

\begin{table*}[t]
\centering
\small
\begin{tabular*}{\textwidth}{@{\extracolsep{\fill}}lccc@{}}
\hline
Comparison & Base [95\% CI] & MLIE-CoT [95\% CI] &
$\Delta$ [95\% CI] \\
\hline
Bagel on GenEval
& 0.8082 [0.7791, 0.8359]
& 0.8566 [0.8279, 0.8840]
& +0.0484 [+0.0430, +0.0555]\\
Bagel on WISE
& 0.7040 [0.6780, 0.7300]
& 0.7600 [0.7330, 0.7860]
& +0.0560 [+0.0400, +0.0720]\\
Flux.2 on GenEval
& .848 [.819, .876]
& .884 [.857, .909]
& +.036 [+.018, +.054] \\
Flux.2 on WISE
& .734 [.708, .760]
& .778 [.752, .803]
& +.044 [+.028, +.060] \\
\hline
\end{tabular*}
\caption{Paired uncertainty and significance results for the two specified
same-prompt comparisons.  Raw/Holm-adjusted randomization $p$-values are
.1100/.1100 for WISE and .0003/.0006 for GenEval.  The WISE interval for the
improvement includes zero and is not significant after correction; the
GenEval improvement remains significant.}
\label{tab:paired_significance}
\end{table*}

\subsection{Paired Confidence Intervals and Significance Tests}
\label{sec:paired_significance}

This section contains Paired Confidence Intervals and Significance Tests on Flux.2-dev model. We quantify uncertainty using the binary outcome for every prompt,
$(\texttt{prompt\_id},\texttt{category},y_i^{\mathrm{base}},
y_i^{\mathrm{MLIE}})$.  For WISE, each of 10,000 bootstrap replicates
samples 1,000 prompt IDs with replacement.  For GenEval, each replicate
instead samples with replacement independently within each of the six
official categories and recomputes Overall as the unweighted mean of the six
resampled category means.  In both cases, the Base and MLIE-CoT outcomes for
a prompt are always selected together.  We report percentile 95\% confidence
intervals from the 2.5th and 97.5th percentiles of the resulting Base,
MLIE-CoT, and paired-difference distributions.

For significance testing, we use a two-sided paired randomization test.
Within each prompt, a random bit determines whether the Base and MLIE-CoT
outcomes are exchanged; Overall is then recomputed using the benchmark's
official aggregation rule.  We use 10,000 random exchanges and the
plus-one Monte Carlo correction.  The two requested Overall tests form one
multiple-testing family, and their $p$-values are adjusted with Holm's
step-down procedure.  All resampling uses seed 20260727.

The paired outcomes contain 40 Base-wrong/MLIE-correct cases and 26
Base-correct/MLIE-wrong cases.  The GenEval comparison uses
\texttt{budget\_0} and the complete \texttt{final} snapshot; it contains 25 and 4 such
discordant cases, respectively.  The latter run uses fixed BAGEL initial
images and FLUX.2 as its only repair executor, so ``FLUX.2-only'' describes
the repair trajectory rather than the initial-image generator.

\end{document}